\documentclass[letterpaper, 10 pt, conference]{ieeeconf}  

\IEEEoverridecommandlockouts                              

\usepackage{cite}
\usepackage{amsmath,amssymb,amsfonts}
\usepackage{algorithmic}
\usepackage{graphicx}
\usepackage{textcomp}
\usepackage{xcolor}
\usepackage{booktabs}
\usepackage{graphicx}
\usepackage{tikz}
\usepackage{subcaption}
\usepackage{multirow} 
\usepackage{comment}
\usepackage{afterpage}

\title{\LARGE \bf
Toward Physics-Aware Deep Learning Architectures for LiDAR Intensity Simulation
}

\author{Vivek Anand$^{1,3,*}$, Bharat Lohani$^{1}$, Gaurav Pandey$^{2}$ and  Rakesh Mishra$^{3}$
\thanks{*Corresponding author: viveka21@iitk.ac.in}
\thanks{$^{1}$Geoinformatics, Department of Civil Engineering, Indian Institute of Technology Kanpur, India. {\tt\small viveka21@iitk.ac.in, blohani@iitk.ac.in}}%
\thanks{$^{2}$Department of Engineering Technology \& Industrial Distribution (ETID),Texas A\&M University, USA. {\tt\small gpandey@tamu.edu}}%
\thanks{$^{3}$Department of Geodesy and Geomatics Engineering, University of New Brunswick, NB, Canada. {\tt\small rakesh.mishra@unb.ca}}%
}

\begin{document}

\maketitle
\thispagestyle{empty}
\pagestyle{empty}

\begin{abstract}
Autonomous vehicles (AVs) heavily rely on LiDAR perception for environment understanding and navigation. LiDAR intensity provides valuable information about the reflected laser signals and plays a crucial role in enhancing the perception capabilities of AVs. However, accurately simulating LiDAR intensity remains a challenge due to the unavailability of material properties of the objects in the environment, and complex interactions between the laser beam and the environment. 
The proposed method aims to improve the accuracy of intensity simulation by incorporating physics-based modalities within the deep learning framework.
One of the key entities that captures the interaction between the laser beam and the objects is the angle of incidence. In this work we demonstrate that the addition of the LiDAR incidence angle as a separate input to the deep neural networks significantly enhances the results. We present a comparative study between two prominent deep learning architectures: U-NET a Convolutional Neural Network (CNN), and Pix2Pix a Generative Adversarial Network (GAN). We implemented these two architectures for the intensity prediction task and used SemanticKITTI and VoxelScape datasets for experiments. The comparative analysis reveals that both architectures benefit from the incidence angle as an additional input. Moreover, the Pix2Pix architecture outperforms U-NET, especially when the incidence angle is incorporated. 
\end{abstract}

\begin{figure*}[h]
\centerline{\includegraphics[width=0.75\textwidth,keepaspectratio]{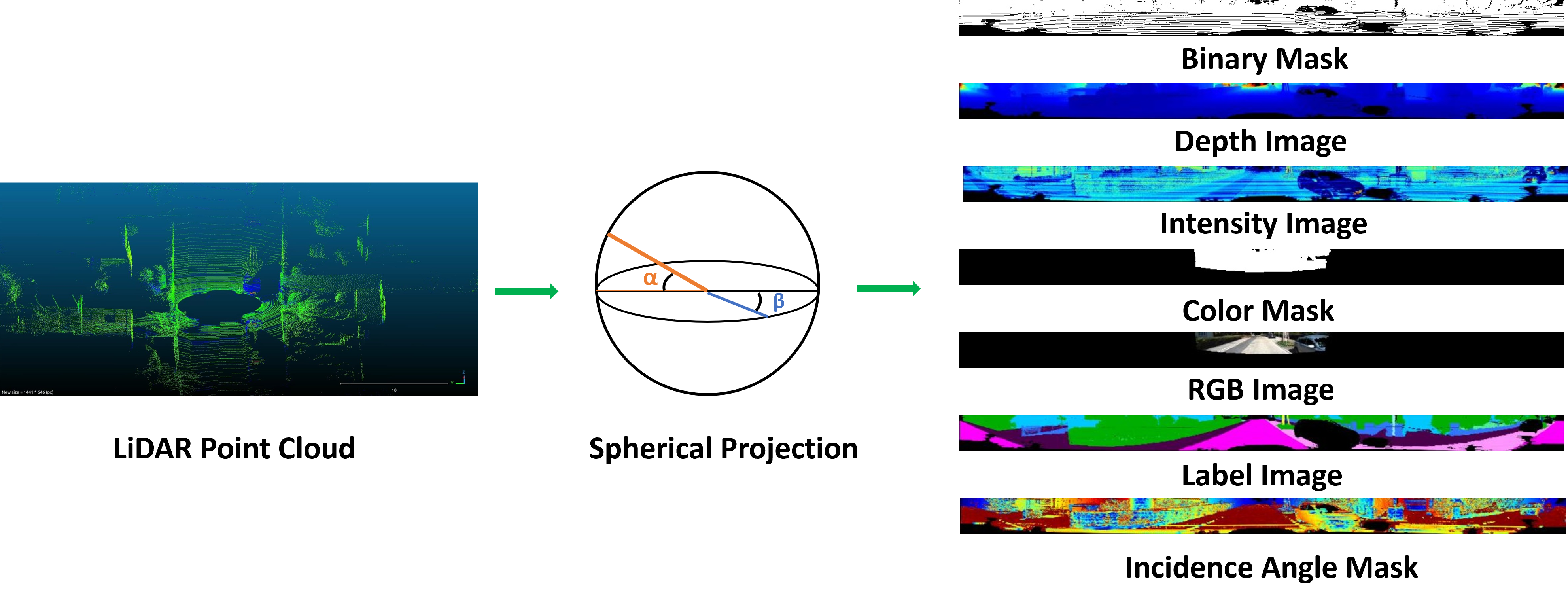}}
\caption{\textbf{Data Preparation:} The LiDAR point cloud is projected on a spherical surface to create LiDAR spherical projection images }
\label{fig-spherical-proj}
\end{figure*}

\section{Introduction}
Autonomous vehicles (AVs) have emerged as a transformative technology that promises to revolutionize transportation systems worldwide \cite{schwarting2018planning}. These vehicles equipped with advanced sensors and intelligent algorithms are capable of navigating and making decisions without human intervention. One of the key sensors used in AVs is LiDAR (Light Detection and Ranging), which plays a crucial role in perceiving the surrounding environment and ensuring safe and efficient autonomous navigation \cite{royo2019overview, li2020LiDAR}. These sensors work on the principle of emitting laser beams and measuring the time it takes for the laser pulses to return after reflecting off objects in the scene \cite{li2020LiDAR}. 
In addition to distance measurements, LiDAR sensors can also capture other attributes such as intensity, which refers to the strength of the reflected laser signal. LiDAR intensity carries valuable information about the characteristics of objects and surfaces, making it an important parameter in various applications, including object recognition, scene understanding, obstacle detection, classification, and semantic segmentation \cite{wang2018pointseg}\cite{wang2019scnet} \cite{yang2018hdnet} \cite{yang2018pixor} \cite{wang2017deep} \cite{meyer2019lasernet} \cite{liang2020rangercnn}.



To train and validate LiDAR perception algorithms, large-scale datasets with ground truth intensity information are required. However, collecting data in the real world is a very expensive and time-consuming task, hence, simulation turns out to be the most promising alternative \cite{yang2021survey}. Currently, LiDAR simulation methods heavily rely on physics-based models and assumptions about material properties and surface interactions. However, accurately simulating LiDAR intensity using these methods is a complex and computationally expensive task as it presents significant challenges due to the dependence on multiple factors, including incidence angle, material properties, and surface interactions among others \cite{incidence2_dai2022LiDAR}\cite{vacek2021learning}. Among these factors, the incidence angle plays a vital role in determining the intensity of the reflected laser signal \cite{incidence2_dai2022LiDAR} \cite{incidence1_tatoglu2012point}. As the angle between the incident laser beam and the surface normal changes, the intensity observed by the LiDAR sensor also changes.


The physics-based approaches and learning-based approaches have quite complimentary challenges. For example, physics-based approaches require that the material property of the environmental objects is known whereas learning-based methods learn the material properties directly from the data but they struggle to capture the intricate interactions between laser pulse and the object surfaces accurately. These limitations hinders the realism and fidelity of simulated LiDAR data, impeding the development and evaluation of perception algorithms for autonomous vehicles. To overcome these challenges, we propose a hybrid learning-based approach that not only leverages the power of deep learning algorithms but also captures the complex interactions between the laser pulse and the object surface more effectively. 
By incorporating the incidence angle into the learning model, it becomes possible to leverage this contextual information and improve the realism and accuracy of LiDAR intensity simulation. 

In this paper, we conduct a comparative study of two different deep learning architectures, namely  U-NET\cite{unet} and Pix2Pix\cite{Pix2pix_gan}, for LiDAR intensity simulation. The novel methodology we propose involves incorporating the incidence angle as an input modality, enhancing both architecture's performance. We conduct extensive experiments using SemanticKITTI data \cite{semantickitti}, which is collected in the real world, and VoxelScape data \cite{voxelscape}, which is simulated using physics-based methods, exploring both established and unexplored methods in the field. The results showcase the importance of the incidence angle in accurately predicting LiDAR intensity and demonstrate that the Pix2Pix architecture, a new addition to the literature for this task, outperforms the existing U-NET approach. Our findings pave the way for more realistic and reliable LiDAR perception in AVs, offering valuable insights into the comparative strengths and weaknesses of different deep learning methodologies in LiDAR intensity simulation.

\begin{figure*}[]
\centerline{\includegraphics[width=0.74\textwidth,keepaspectratio]{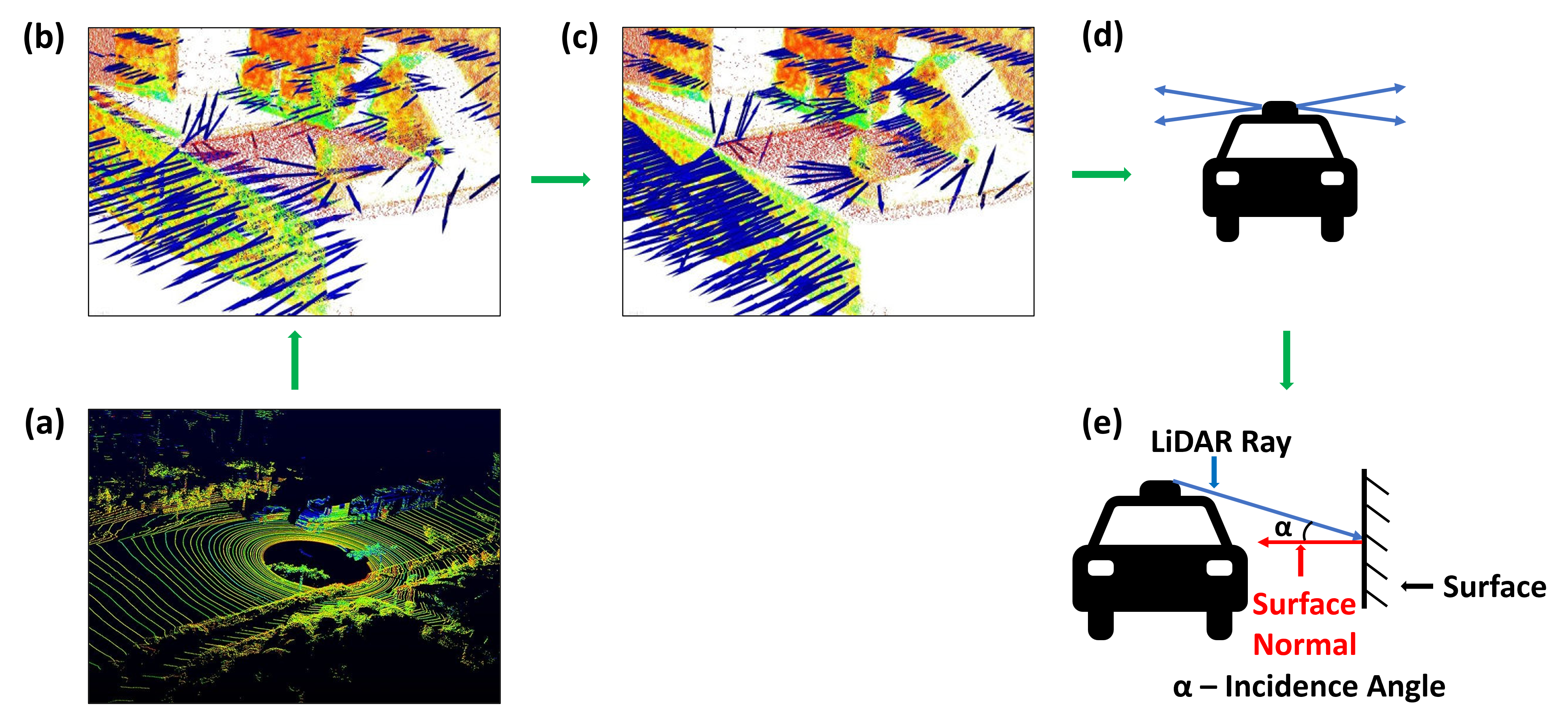}}
\caption{\textbf{Incidence Angle Calculation:}\textbf{(a)} LiDAR point cloud \textbf{(b)} Estimating surface normal \textbf{(c)} Orienting surface normal towards the sensor \textbf{(d)} Computing the direction vector of LiDAR rays \textbf{(e)} Computing the dot product between the direction and normal vectors of the point to get the incidence angle}
\label{fig-incidence-angle}
\end{figure*}
\section{Related Work}

\subsection{Simulation for Autonomous Vehicles}

Simulation plays a vital role in the development and evaluation of autonomous vehicles, particularly in mimicking real-world driving conditions, enabling comprehensive testing of perception algorithms and the assessment of their performance\cite{yang2021survey}. It provides a controlled and repeatable environment for algorithm development and evaluation, allowing researchers to precisely control variables and compare different approaches. Additionally, simulation overcomes limitations in data availability of different sensors used in autonomous vehicles including LiDAR by generating large-scale datasets with diverse scenarios, supporting algorithm training and validation \cite{geiger2012we} \cite{caesar2020nuscenes} \cite{sun2020scalability}. 

\subsection{Physics-based LiDAR Simulation}

Physics-based LiDAR simulation is a widely used approach for generating synthetic LiDAR data to test and evaluate autonomous driving systems\cite{yang2021survey}. This method involves modeling the physical processes of laser beam emission, propagation, and interaction with objects in the environment\cite{elmquist2020methods}. By considering factors such as surface reflectance, material properties, and light scattering effects, physics-based simulations aim to accurately replicate the behavior of real-world LiDAR sensors. However, simulating LiDAR intensity poses significant challenges due to the complex nature of interactions between the laser beam and the objects in the scene\cite{vacek2021learning}. Precisely modeling these interactions and accounting for various environmental factors are extremely difficult, computationally expensive, and may result in inaccuracies when compared to real-world LiDAR intensity measurements.

\subsection{Learning-based LiDAR Simulation}

Learning-based approaches are emerging as a promising alternative for simulating LiDAR sensors. Instead of relying on complex physics-based models, these methods leverage the power of deep learning algorithms to learn the complexity of the factors involved from the data itself \cite{marcus2022lightweight}. Learning-based LiDAR simulation offers several advantages, including improved computational efficiency, flexibility in handling different environmental conditions, and the potential to capture subtle nuances that may be challenging to model using physics-based methods. By training neural networks on large datasets of real-world LiDAR measurements, these models can capture the underlying patterns and dependencies between input features and intensity values. Recently, a generative adversarial network (GAN) for intensity rendering in LiDAR simulation was introduced by \cite{mok2021simulated} where they transform the 3D point cloud to a 2D spherical image and project the intensity data on the pixel values to learn intensity based on unpaired real and synthetic data. Some of the recent works (\cite{nakashima2021learning}, \cite{caccia2019deep} and \cite{manivasagam2020LiDARsim}) have also tried to predict the reflected signal (also called ray-drop) and estimate the intensity from the returned signal by applying some detection threshold and hence do not predict the intensity values directly. \cite{guillard2022learning} aims to simulate an enhanced point cloud with ray drops and intensity values but they project the LiDAR data and the corresponding intensity values on the RGB image space and the model learns and generates the value in the RGB image space itself and hence the intensity values can not be generated for all the points in the LiDAR data. \cite{vacek2021learning} utilize a fully supervised training approach with Convolutional Neural Networks (CNNs) using LiDAR data and its derived modalities like range, RGB, and label image to predict intensity values. They employ their trained model to generate synthetic LiDAR data with intensity by applying it to ray-casted point clouds generated in the gaming engine.

In this paper, we included the incidence angle in conjunction with range, RGB values, and semantic labels, building upon the work of  \cite{vacek2021learning} and extended the comparative study to include Generative Adversarial Network (GAN) by implementing Pix2Pix architecture. By including incidence angle as an input, learning-based models can better account for the influence of surface orientation on LiDAR intensity, leading to more accurate simulations. The rest of the paper is organized as follows: section \ref{sec:method} describes the proposed methodology, section \ref{sec:results} presents results on open source datasets and provides discussions about the findings, and finally section \ref{sec:conclusions} provides concluding remarks.

\begin{figure*}[]
  \centering
  \includegraphics[width=0.74\textwidth,keepaspectratio]{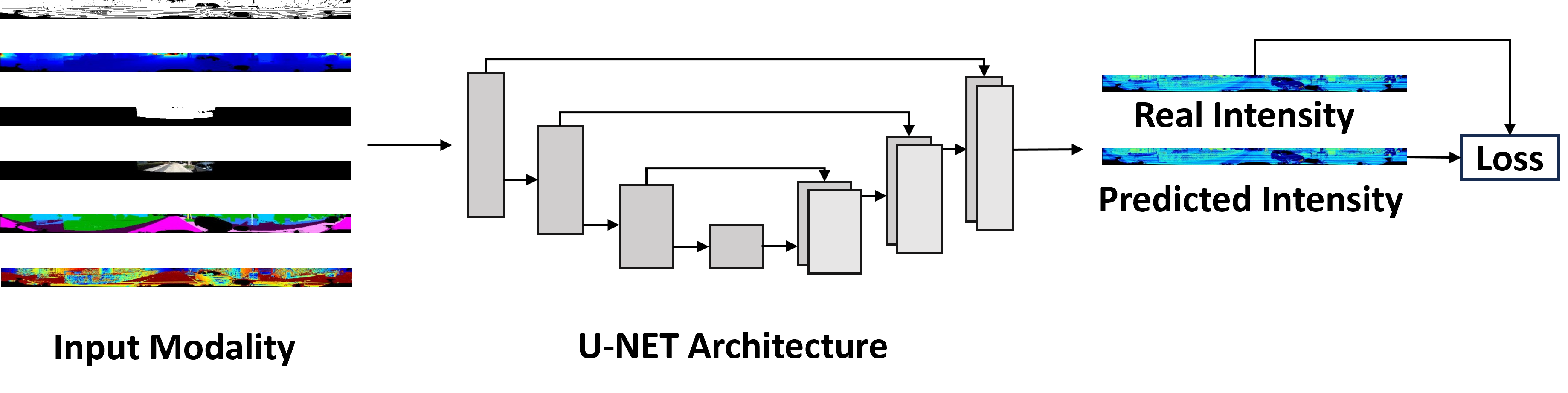}
  
  \vspace{1em}  
  
  \includegraphics[width=0.74\textwidth,keepaspectratio]{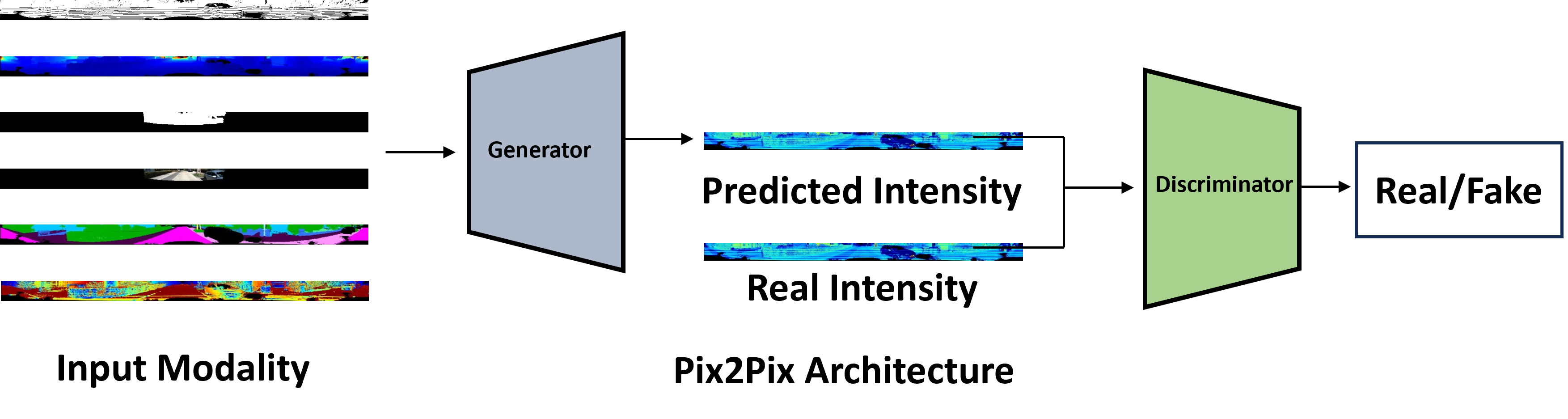}
  
  \caption{\textbf{Training pipeline:} LiDAR Spherical images are fed into the architecture to predict LiDAR intensity.}
  \label{fig:training_pipeline}
\end{figure*}

\section{Method} \label{sec:method}

We propose to develop a hybrid learning-based approach that takes into account the physical interactions between the laser and the environment objects. We propose to use the physics-based models combined with the novel deep learning architectures to improve the realism and accuracy of LiDAR intensity. To achieve this, we employ a spherical projection technique \cite{sph_proj_li2016vehicle} to transform the 3D LiDAR points into a structured 2D image representation (Fig. \ref{fig-spherical-proj}). This involves mapping the LiDAR points onto a virtual spherical surface centered at the LiDAR sensor origin. Each LiDAR point's coordinates are converted into spherical coordinates, including azimuth and elevation angles, which determine their positions on the spherical surface. This spherical projection preserves the geometric relationships and distances between LiDAR points, enabling the capture of spatial information in a structured format. The resulting LiDAR spherical projection image serves as a foundation for deriving various modalities used in the study.
 
The derived modalities from LiDAR and other complimentary sensor (e.g. camera) data include depth image, logical binary Mask, segmented spherical image, RGB spherical image, LiDAR color mask, and incidence angle mask. The LiDAR depth image captures the distance information of LiDAR points from the sensor, providing a depth-based representation. The LiDAR logical binary mask indicates the presence or absence of returned LiDAR rays. The LiDAR segmented image represents the semantic segmentation of LiDAR points based on their assigned labels. The LiDAR RGB spherical image projects RGB values onto LiDAR points from corresponding camera images, providing color information. The LiDAR Color Mask is a binary mask indicating the presence or absence of RGB values for each LiDAR point. The incidence angle mask represents the angle of incidence of the laser beam at the given 3D coordinate point. The integration of these diverse modalities enhances the accuracy and realism of LiDAR intensity prediction.

Calculation of the incidence angle for each point in the LiDAR point cloud is a critical component of our proposed methodology, the entire process is shown in Fig \ref{fig-incidence-angle}. The incidence angle represents the angle between the LiDAR ray direction and the surface normal at a given point. To determine the local orientation of the underlying surface, we estimate the surface normal at each point in the point cloud. The estimated normal is then oriented towards the sensor origin to ensure consistent directionality. By considering the Euclidean distance from the sensor origin to each point, we calculate the radial position of the point. Next, we compute unit direction vectors from the sensor origin to each point by normalizing the direction vectors. The dot product between the direction vectors and the normal vector for each point is then calculated to determine the angle between them. To ensure the correct orientation of the normal vectors towards the sensor, we check if the dot product is negative for any point. If a negative dot product is found, indicating that the normal vector points away from the sensor, the normal vector is flipped to ensure it points toward the sensor. The dot product is recalculated after flipping the normal vectors. The incidence angle for each point is obtained by taking the arccosine of the dot product. These calculated incidence angles serve as valuable information for understanding the interaction between the LiDAR sensor and the surfaces. 



\subsection{Network Architecture}

In this study, we investigate two different categories of deep learning architectures for the task of LiDAR intensity prediction, namely U-NET and Pix2Pix, each offering distinct characteristics and capabilities.

\begin{table*}[ht]
\centering
\caption{Comparison of MSE Loss  [D - Depth, I - Incidence Angle, L - Label, RGB - Color]}
\label{tab:mse-loss}
\begin{tabular}{llrrrrrrrr}
\toprule
Architecture & Dataset & D & D+I & D+L & D+L+I & D+RGB & D+RGB+I & D+RGB+L & D+RGB+L+I \\
\midrule
U-NET & SemanticKiTTI & 0.353 & 0.349 & 0.339 & 0.321 & 0.317 & 0.301 & 0.298 & 0.234 \\
& VoxelScape & 0.373 & 0.291 & 0.341 & 0.263 & - & - & - & - \\
\midrule
Pix2Pix & SemanticKiTTI & 0.334 & 0.325 & 0.319 & 0.299 & 0.295 & 0.272 & 0.253 & \textcolor{blue}{\textbf{0.201}} \\
& VoxelScape & 0.247 & 0.192 & 0.211 & \textcolor{blue}{\textbf{0.182}} & - & - & - & - \\
\bottomrule
\end{tabular}
\end{table*}

\subsubsection{U-NET Architecture}
The widely used UNET architecture is employed in the first part of the study, popular for semantic segmentation tasks and recognized for its ability to capture fine-grained spatial details. To adapt UNET for this specific task, the last layer is removed, allowing the direct output of the predicted LiDAR intensity and bypassing non-linear transformations. Multiple modalities are fed into the network as separate channels which are described earlier along with the novel addition of LiDAR Incidence Angle Mask. This multi-channel input enables the network to utilize the diverse information captured by each modality during the prediction process.
The output of the network is the predicted LiDAR intensity, which aims to accurately represent the intensity values associated with each corresponding point in the input LiDAR point cloud. 
The loss function used is an extended version of L2 loss called masked L2 loss. The masked L2 loss incorporates a binary mask (B) as an additional input channel in the loss calculation, enabling the model to focus on the returned scan points during training.
\begin{equation}
L = \frac{1}{n} \sum_{i,j} (I_{i,j} - \hat{I}_{i,j})^2 \cdot B_{i,j} 
\end{equation}

It measures the difference between the real intensity values(I) and predicted intensity values$(\hat{I}$ ) for each pixel(i,j) in the spherical image. The binary mask(B) indicates the presence of returned scan points, and the loss is calculated by summing the squared errors and normalizing by the total number of successful rays. The loss quantifies the accuracy of intensity predictions in the image. To optimize the model, we employed the Adam algorithm with a learning rate of 0.003 and weight decay of 0.001 building upon the work of \cite{vacek2021learning}. The experiments were performed on SemanticKITTI and VoxelScape datasets respectively and the training dataset was divided into 7500 training frames and 2500 validation frames. To evaluate the performance of the LiDAR intensity prediction model, test sets consisting of 2500 frames each were used. By using these test sets, we assessed the model's accuracy and its ability to generalize to different datasets and scenarios.

\begin{figure*}[]
\centerline{\includegraphics[width=0.78\textwidth,keepaspectratio]{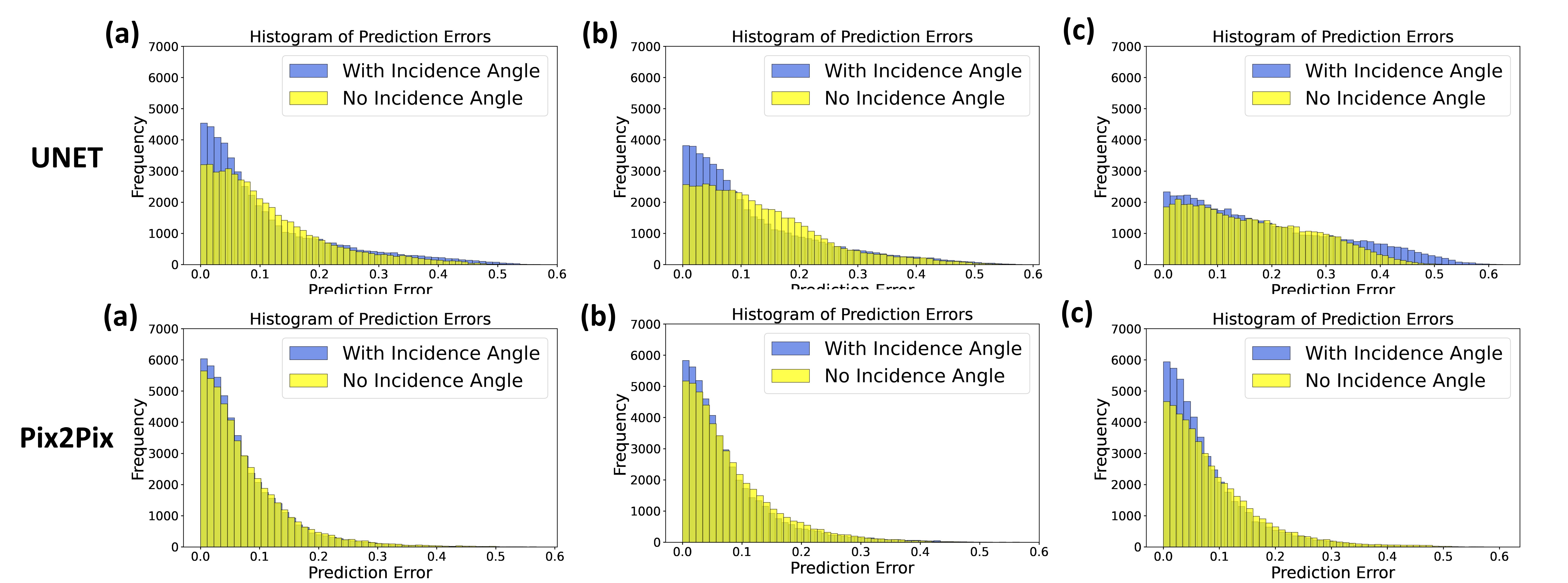}}
\caption{\textbf{Error Histogram}: SemanticKITTI Data - Input combinations: (a) Depth + RGB + Label (b) Depth + RGB (c) Depth}
\label{Error Hist - Kitti}
\end{figure*}

\

\subsubsection{Pix2Pix Architecture}
In the second part of the study, we introduce the novel use of the Pix2Pix architecture for this specific task. The Pix2Pix architecture, unlike U-NET, leverages a conditional generative adversarial network (cGAN) that consists of a generator and a discriminator. The generator aims to transform an input image into the desired output (LiDAR intensity in this case), while the discriminator's task is to distinguish between real and generated images. For this study, the generator part of the Pix2Pix model was adapted to take in the same multi-channel inputs as the U-NET. The generator part employs a series of encoder-decoder layers to capture the relationship between the different input modalities and the LiDAR intensity. 
The discriminator which employs a PatchGAN structure \cite{pix2pix}, in parallel, assesses the authenticity of the generated LiDAR intensity image. By being trained to recognize real-intensity images, it provides feedback to the generator, guiding it to produce more realistic predictions.

The loss function used in the Pix2Pix model is a combination of a conditional adversarial loss and an L1 loss \cite{Pix2pix_gan}, as defined below: 
\begin{equation}
L = \lambda \cdot \text{L1 Loss} + \text{Adversarial Loss}    
\end{equation}

Here, the L1 Loss measures the absolute difference between the real intensity values (I) and predicted intensity values ($\hat{I}$) for each pixel (i,j) in the spherical image, and the Adversarial Loss quantifies how well the generator fools the discriminator.

The Pix2Pix architecture was optimized using the Adam algorithm with a learning rate of 0.0002, $\lambda$ value of 100, and a  gradient penalty term with a coefficient of 10. Similar to the U-NET model, the experiments were carried out on SemanticKITTI and VoxelScape datasets. The division of training and validation frames and the evaluation metrics were kept consistent with the U-NET model to ensure a fair comparison.

\begin{figure*}[]
\centerline{\includegraphics[width=0.6\textwidth,keepaspectratio]{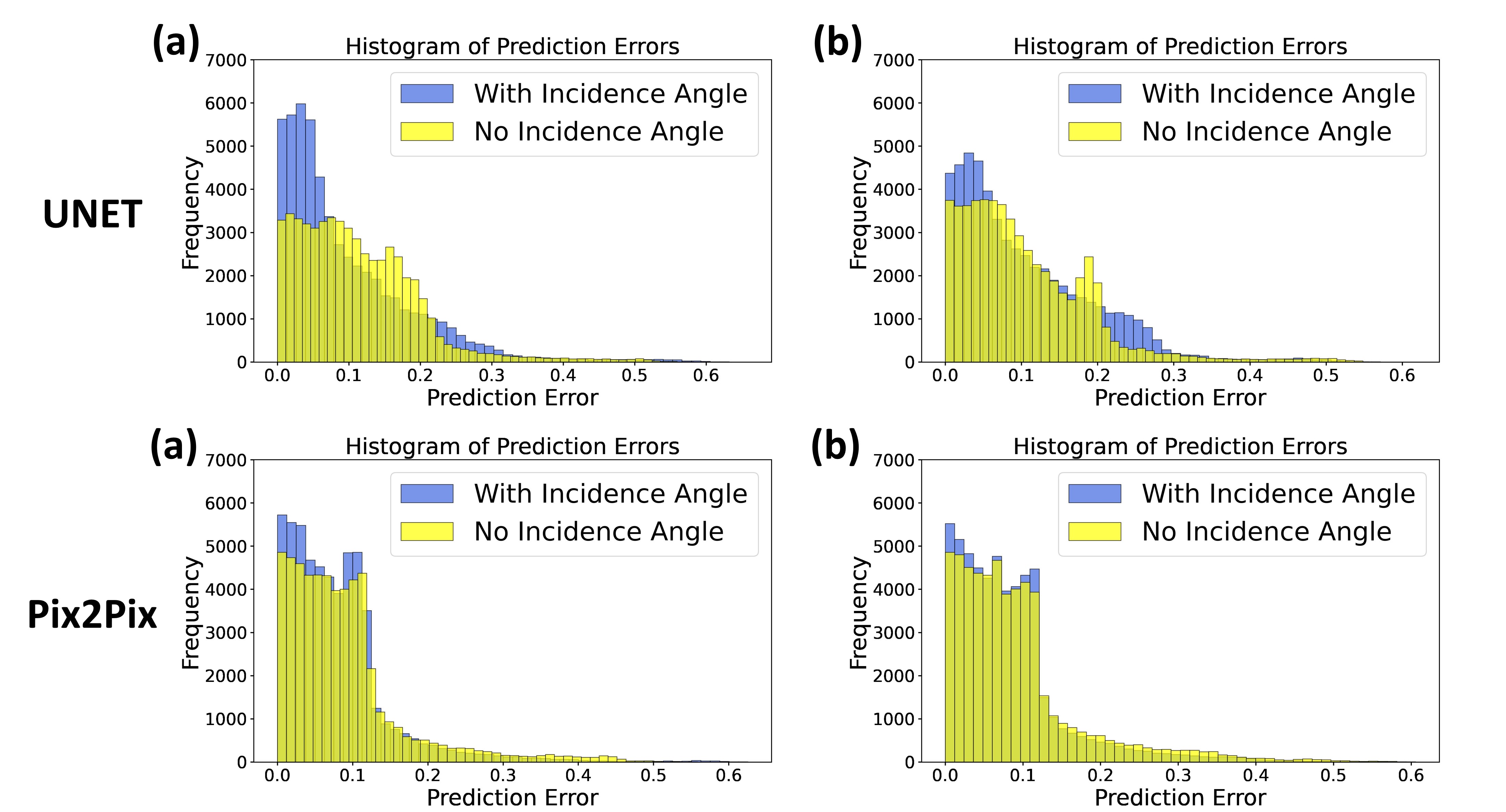}}
 \caption{\textbf{Error Histogram}: VoxelScape Data - Input combinations: (a) Depth + Label (b) Depth} 
  \label{Error Hist - Voxel}
\end{figure*}

\section{Results and Discussion} \label{sec:results}

The LiDAR intensity prediction experiments were conducted using two datasets: SemanticKITTI\cite{semantickitti} and VoxelScape\cite{voxelscape}. To assess the performance of different input modalities, the mean squared error (MSE) loss was calculated for both U-NET and Pix2Pix GAN architectures on the test set. Further, error histograms and heatmaps were generated to analyze the prediction errors and their spatial distribution. The experiments were performed on SemanticKitti and Voxelscape datasets, respectively, and the training dataset was divided into 7500 training frames and 2500 validation frames. To evaluate the performance of the
Lidar intensity prediction model, test sets consisting of 2500 frames each were used.

\subsection{Mean Squared Error (MSE)}

In ablation studies involving both the SemanticKITTI and VoxelScape datasets, the performances of U-NET and Pix2Pix architectures were compared using various input modalities. The results show that the Pix2Pix architecture outperforms U-NET in all instances. Importantly, the addition of the incidence angle across both architectures consistently led to improved performance, as highlighted in Table \ref{tab:mse-loss}. This highlights the significance of incorporating the incidence angle as an informative input modality for LiDAR intensity prediction. The VoxelScape data does not contain RGB images; hence, the experiment results involving RGB information for Voxelscape are not present in Table \ref{tab:mse-loss}. 

\subsection{Error Histogram}

To further analyze the prediction errors between the ground truth and predicted intensity values, error histograms for each frame from the SemanticKITTI and VoxelScape datasets were generated. These histograms visualize the distribution of prediction errors across the intensity range. It was observed that in the results from Pix2Pix architecture, the error distribution is concentrated more around zero error compared to the U-Net architecture, and when the incidence angle was included it was further improved indicating a better alignment between predicted and ground truth intensity values (Fig \ref{Error Hist - Kitti} and Fig \ref{Error Hist - Voxel}). This supports the notion that the incorporation of the incidence angle enhances the accuracy of the LiDAR intensity prediction.

\begin{center}
\begin{figure*}[h]
\centering
\renewcommand\thesubfigure{\roman{subfigure}}
\begin{subfigure}{0.48\textwidth}
  \centering
 \includegraphics[width=\textwidth, keepaspectratio]{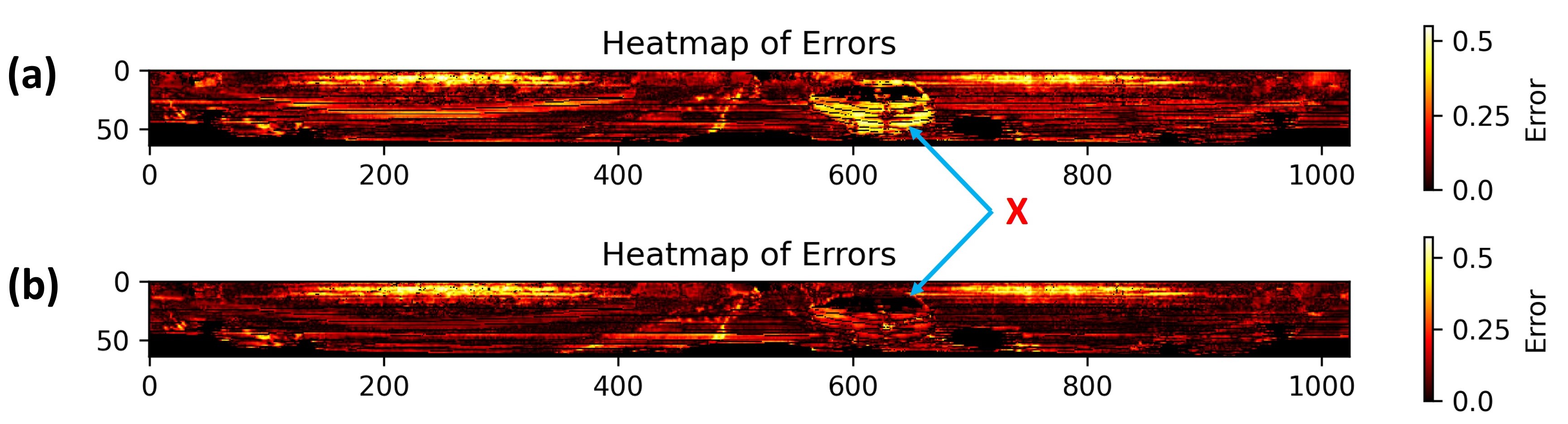}
  \caption{Pix2Pix}
  \label{fig:heatmap}
\end{subfigure}
\hfill
\begin{subfigure}{0.48\textwidth}
  \centering
  \includegraphics[width=\textwidth, keepaspectratio]{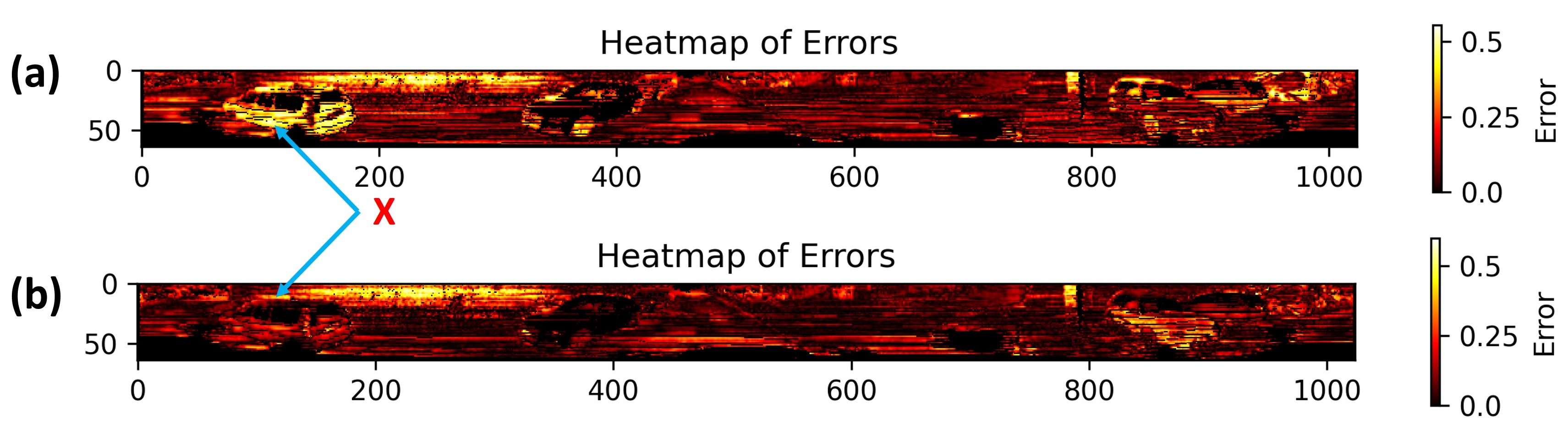}
 \caption{U-NET}
  \label{fig:b}
\end{subfigure}
\caption{\textbf{Error Heatmap:} Error heatmap a) With Incidence Angle b) Without Incidence Angle }
\label{fig:heatmap}
\end{figure*}

\begin{figure*}[h]
\centering
\renewcommand\thesubfigure{\roman{subfigure}}
\begin{subfigure}{0.48\textwidth}
  \centering
  \includegraphics[width=\textwidth, keepaspectratio]{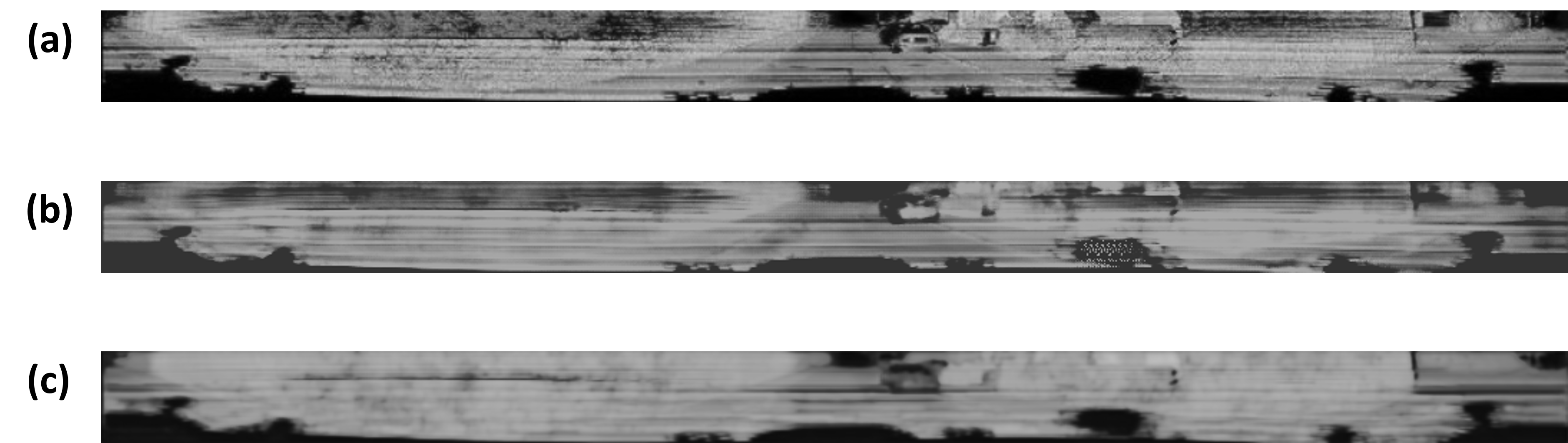}
  \caption{SemanticKITTI}
  \label{fig:Generated Images}
\end{subfigure}
\hfill
\begin{subfigure}{0.48\textwidth}
  \centering
  \includegraphics[width=\textwidth, keepaspectratio]{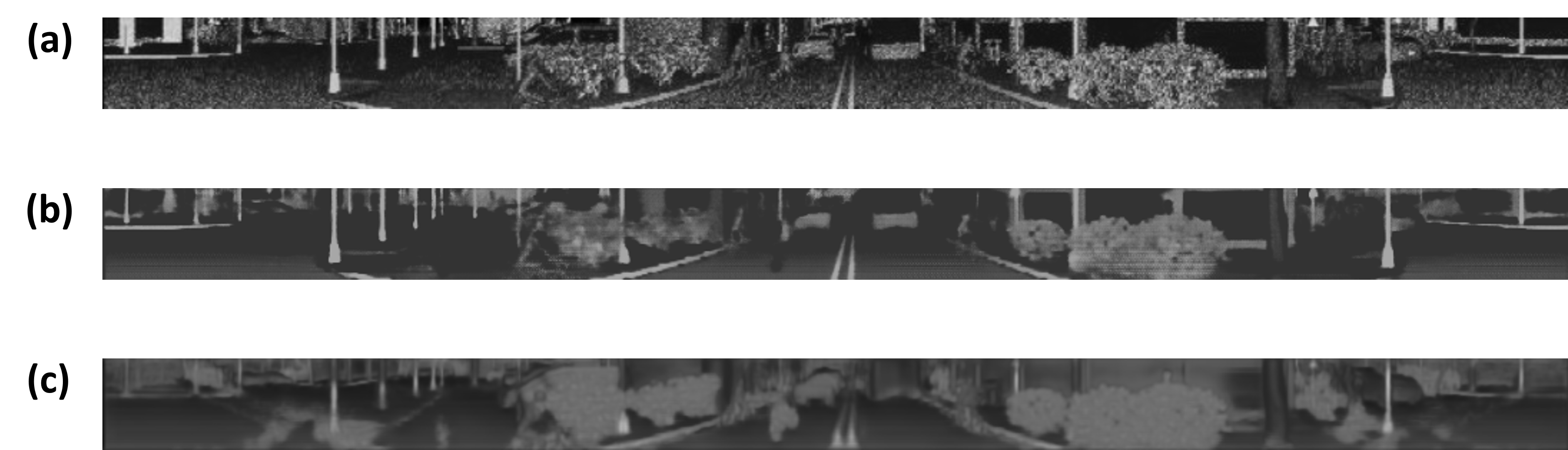}
  \caption{VoxelScape}
  \label{fig:b}
\end{subfigure}
\caption{\textbf{Generated Images:} a) Reference LiDAR intensity image b) Pix2Pix architecture c) U-NET architecture}
\label{fig:Generated Images}
\end{figure*}
\end{center}

\subsection{Error Heatmaps}
Furthermore, error heatmaps were generated to visualize the spatial distribution of prediction errors and assess the impact of incorporating incidence angle on LiDAR intensity prediction.  
Upon examining the heatmaps, a distinct pattern emerges, showcasing localized improvements in predicted intensity, as shown in Fig \ref{fig:heatmap}, the error is significantly reduced for the cars present in the image (marked as X). Integrating the incidence angle results in localized refinements, reducing prediction errors in distinct regions, thereby enhancing the spatial consistency and accuracy of the intensity prediction. This phenomenon can be attributed to the significant influence that incidence angle has on the intensity of LiDAR reflections. By incorporating this crucial factor, the model gains a deeper understanding of the underlying knowledge of the interactions between the LiDAR rays and the surface, capturing the variations in reflectivity based on surface orientation, resulting in more precise and contextually aware intensity predictions.

\subsection{Qualitative Analysis}

For a visual assessment, we looked closely at the images generated by both U-NET and Pix2Pix and compared them to the reference images from the SemanticKITTI and VoxelScape datasets as shown in Fig \ref{fig:Generated Images}. In this comparison, it was clear that the Pix2Pix images are closer to the reference images than the U-NET ones. U-NET produces blurry output as also mentioned by \cite{pix2pix} and Pix2Pix did a far better job at capturing the details present in the scene. 

While both the architectures share U-NET as the core architecture, Pix2Pix leverages a Generative Adversarial Network (GAN) training framework, setting it apart from U-NET's reliance solely on pixel-wise loss (L2 loss). This distinction in loss objectives explains the Pix2Pix's edge. U-NET minimizes pixel-wise errors, while Pix2Pix's adversarial loss encourages the model to match ground truth and generate outputs that deceive a discriminator network, potentially leading to more refined and realistic outcomes, hence outlining the importance of generative adversarial network (GAN) in simulating LiDAR data.

\section{Conclusion} \label{sec:conclusions}
In this research paper, we presented a comprehensive comparative study involving two different deep learning architectures U-NET and Pix2Pix for LiDAR intensity simulation. The core novelty of this approach lies in incorporating the incidence angle as an additional input modality, enhancing both the accuracy and context awareness of LiDAR intensity predictions. Through extensive experiments using both SemanticKITTI and VoxelScape data, we demonstrated the effectiveness of the U-NET and Pix-2Pix architecture with the inclusion of the incidence angle. We showed that this combination leads to improvements in prediction performance. Further, our implementation of the Pix2Pix architecture in this context, a novel approach in the literature, exhibited even better results, thus underscoring the potential of generative adversarial networks in LiDAR intensity simulation.
Our findings consistently revealed that the integration of the incidence angle, along with other modalities led to superior performance across both architectures. This success not only addresses the inherent challenges posed by the complex nature of intensity simulation but also reflects a significant step forward in narrowing the gap between simulated and real-world LiDAR data.

The successful integration of the incidence angle and the comparative analysis of U-NET and Pix2Pix represent innovative and meaningful contributions to the field of LiDAR intensity simulation. These advancements have tangible implications for various applications, including autonomous driving, robotics, and environmental mapping, where accurate LiDAR intensity plays a crucial role in perception, object recognition, and scene understanding. For future research, we will try to explore other factors, such as material properties, surface reflectance, and the development of network architectures and loss functions designed specifically for LiDAR intensity simulation

\section*{Acknowledgment}

This work was supported by the Mitacs Globalink Research Award. We are also grateful to the University of New Brunswick, Canada, for providing access to \textit{Compute Canada}, the national high-performance compute (HPC) system, for our experiments.

\bibliographystyle{ieeetr}
\bibliography{reference}
\end{document}